\documentclass{article}
\usepackage{ijcai16}

\usepackage{url}
\usepackage{hyperref}
\usepackage[svgnames]{xcolor}

\usepackage[authoryear,round]{natbib}

\usepackage{times}

\usepackage{latexsym} 

\usepackage{graphicx}
\usepackage[T1]{fontenc}

\newcommand{\nec}{\Box}

\title{Universal Reasoning, Rational Argumentation and Human-Machine Interaction}
\author{Christoph Benzm\"uller \\ University of Luxemburg \& Freie Universit\"at Berlin\\ christoph.benzmueller@uni.lu | c.benzmueller@fu-berlin.de}

\begin{document}
\maketitle

\begin{abstract}
Classical higher-order logic, when utilized as a meta-logic in which various other (classical and non-classical) logics can be shallowly embedded, is well suited for realising a universal logic reasoning approach.  Universal logic reasoning in turn, as envisioned already by Leibniz, may support the rigorous formalisation and deep logical analysis of rational arguments within machines. A respective universal logic reasoning framework is described and a range of exemplary applications are discussed. In the future, universal logic reasoning in combination with appropriate, controlled forms of rational argumentation may serve as a communication layer between humans and intelligent machines. 
 \end{abstract}




\section{Rational Argumentation -- Communication Interface between Humans and Machines}
The ambition to understand, model and implement \emph{rational argumentation} and \emph{universal logical reasoning} independent of the human brain has a long tradition in the history of humankind. It reaches back at least to the prominent study of syllogistic arguments by Aristoteles.  Today, with the event of increasingly intelligent computer technology, the question is more topical than ever: if humans and intelligent machines are supposed to amicably coexists, interact and collaborate, appropriate forms of communication between them are required. For example, machines should be able to depict, assess and defend their (options for) actions and decisions in a form that is accessible to human understanding and judgement. This will be crucial for achieving a reconcilable and socially accepted integration of intelligent machines into everyday (human) life. The communication means between machines and humans should ideally be based on human-level, rational argumentation, which since ages forms the fundament of our social, juridical and scientific processes. Current developments in artificial intelligence, in contrast, put a strong focus on statistical information, machine learning and subsymbolic representations, all of which are rather detached from human-level rational explanation, understanding and judgement. The challenge thus is to complement and enhance these human-unfriendly forms of reasoning and knowledge representation in todays artificial intelligence systems with suitable explanations amenable to human cognition, that is, rational arguments. \emph{Via exchange of rational arguments at human-intuitive level the much needed mutual understanding and acceptance between humans and intelligent machines can eventually be guaranteed.} This is particularly relevant for the assessment of machine actions in terms of legal, ethical, moral, social and cultural norms purported by humans.
But what formalisms are available that could serve as a most general basis for the modeling of human-level rational arguments in machines? 

\section{Leibniz' Vision}
The quest for a most general framework supporting universal reasoning and rational argumentation is very prominently represented in the works of Gottfried Wilhelm Leibniz (1646-1716). He envisioned a \textit{scientia generalis} founded on a  \textit{characteristica universalis}, that is, a most universal formal language in which all knowledge (and all arguments) about the world and the sciences can be encoded. This universal logic framework should, so Leibniz, be complemented  with a \textit{calculus ratiocinator}, an associated, most general formal calculus in which the truth of sentences expressed in the characteristica universalis should be mechanically assessable by computation.\footnote{Leibniz characteristica universalis and calculus ratiocinator are prominently discussed 
in the numerous philosophy books and papers. Recommended texts include \citet{Lenzen04} and \citet{peckhaus04:_calcul}.}
Leibniz' envisioned, for example,  that disputes between philosophers could be resolved by formalisation and computation:
\textit{``If this is done, whenever controversies arise, there will be no more need for arguing among two philosophers than among two mathematicians. For it will suffice to take the pens into the hand and to sit down by the abacus, saying to each other (and if they wish also to a friend called for help): Let us calculate.''} (Leibniz 1690, translation by \citet[p. 1]{Lenzen04}).\footnote{\textit{Quo facto, quando orientur controversiae, non magis disputatione opus erit inter duos philosophos, quam inter duos Computistas. Sufficiet enim calamos in manus sumere sedereque ad abacos, et sibi mutuo (accito si placet amico) dicere: calculemus.} (Leibniz 1684; cf. \citet[p. 200]{gerhardt62:_schrif_g}).}

Leibniz' visionary proposal, which became famous under the slogan \emph{Calculemus!: ``Let us Calculate.''}, is very ambitious and far reaching:
\textit{``If we had it [a characteristica universalis], we should be able to reason in metaphysics and morals in much the same way as in geometry and analysis.''} (Leibniz 1677, Leter to Gallois; translation by Russell).\footnote{\textit{Car si nous l'avions telle que je la concois, nous pourrions raisonner en metaphysique et en morale au peu pres comme en Geometrie et en Analyse, \ldots} 
(Leibniz, Leter to Gallois, 1677; cf. \citet[p. 21]{gerhardt62:_schrif_g}).}

From the perspective of the initially depicted challenge, an obvious proposal hence is to extend and adapt Leibniz proposal in particular to disputes (and interaction in general) between humans and intelligent machines.  But how realistic is a characteristica universalis and an associated calculus ratiocinator? 
What has modern logic to offer?



\section{Zoo of Logical Formalisms}
A quick study of the survey literature on logical formalisms\footnote{See for example various handbooks on logical formalisms such as \citet{gabbay14:_handb_histor_logic,benthem11:_handb_logic_languag_secon_edition,gabbay14:_handb_philos_logic,abramsky01:_handb_logic_comput_scien,d.m.98:_handb_logic_artif_intel_logic_progr,blackburn06:_handb_modal_logic}.}
suggest that quite the opposite to Leibniz' dream has become todays reality. Instead of a characteristica universalis, a most general universal formalism supporting rigorous formalisations across all scientific disciplines, we are today actually  facing a very \emph{rich and heterogenous zoo of different logical systems}. Their development is typically motivated by e.g. different practical applications,  different theoretical properties, different practical expressivity, or different schools of origin.
Some exemplary species in the logic zoo are briefly outlined: \\


\noindent \fbox{
\begin{minipage}{.95\columnwidth} \small
On the side of classical logics there are propositional, first-order, second-order and full higher-order logics. 
When rejecting certain basic assumptions, such as the law of excluded middle, we arrive at intuitionistic and constructive logics, where we may again 
distinguish propositional, first-order and higher-order variants. 
Higher-order logics, classical or non-classical, are typically typed (to rule out paradoxes and inconsistencies) and different type systems have been developed. This brings us in the area of type theories (some proof assistants may (additionally) apply the propositions as types paradigm and encode theorems as types and proofs as terms.)
Then there are numerous, so called non-classical logics, including modal logics and conditional logics, logics of time and space, provability logics, multivalued logics, free logics, to name just a few examples.  
Deferring the explosion principle (from falsity anything follows) we arrive at paraconsistent logics.  Moreover, various special purpose logics, e.g. seperation logics and security logics, have recently been developed for particular applications.
Many of the mentioned logic species, e.g. modal logics, have again a wide range of subspecies (e.g. logics K, KB, KT, S4, S5 and different domain conditions for quantified modal logics, etc.). 
And, to further complicate matters, certain practical applications may even require flexible combinations of logics.
\end{minipage}
}

\vskip1em

Many of the above logic formalisms have their origin in philosophy and they have then been picked up and further developed in e.g. computer science, artificial intelligence, computational linguistics and mathematics. 
Instead of converging towards a single superior logic, the logic zoo is obviously further expanding, eventually even at accelerated pace. As a consequence, the unified vision of Leibniz seems further remote from todays reality than ever before. 

However, there are also some promising initiatives to counteract these diverging developments. Attempts at unifying approaches to logic include categorial logic \citep{lambek86:_introd_higher_order_categ_logic,jacobs99:_categ_logic_type_theor}, algebraic logic~\citep{andreka17:_univer_algeb_logic} and coalgebraic logic \citep{moss99:_coalg_logic,rutten00:_univer}. Generally, these approaches have a strong emphasis on theory. However, some promising practical work has recently been reported utilizing the algebraic logic approach~\citep{DBLP:conf/icfem/GuttmannSW11,foster15:_fine_struc_regul_algeb}. 

This paper defends another alternative at universal logical reasoning. This approach has a very pragmatic motivation, foremost reuse of tools, simplicity  and elegance. It utilises classical higher-order logic\footnote{Classical higher-order logic has its roots in the logic of Frege's Begriffsschrift \citep{frege79:_begrif}. However, the version of HOL as used here is a (simply) typed logic of functions, which has been proposed by \citet{Church40}.  It provides lambda-notation, as an elegant and useful means to denote unnamed functions, predicates and sets (by their characteristic functions).  Types in HOL eliminate paradoxes and inconsistencies: e.g. the well known Russel paradox (set of sets which do not contains themselves), which can be formalized in Frege's logic, cannot be represented in HOL due to type constraints. More information on HOL and its automation is provided by \citet{B5}.} (HOL) as a unifying meta-logic in which (the syntax and semantics) of varying other logics can be explicitly modeled and flexibly combined. Off-the-shelf higher-order interactive and automated theorem provers can then be employed to reason about and within the shallowly embedded logics. This way Leibniz vision can (at least partially) be realised.

However, note the difference to Leibniz original idea: Instead of a single, universal logic formalism, the \emph{semantical embedding approach} supports different competing object logics from the logic zoo. They are selected according to the specific requirements of particular applications, and, if needed, they  may be combined. Only at meta-level a single, unifying logic is provided: HOL (or any richer logic incorporating HOL, provided that strong automation tools for it exist). By unfolding the object logic encodings, problem representations are uniformly mapped to HOL. This way Leibniz vision is realized in an indirect way: \emph{universal logical reasoning is established at the meta-level in HOL}.



\section{HOL as Unifying Meta-Logic}
Translations between logic formalisms are not new. For example, by suitably encoding Kripke style semantics (possible world semantics) many propositional modal logics (PMLs) can be translated to classical first-order logic (FOL)~\citep{DBLP:books/el/RV01/OhlbachNRG01,DBLP:conf/birthday/SchmidtH13}. Modulo such transformations, a range of PMLs can thus be uniformly characterized as particular fragments of FOL. Moreover, with the help of respective (external) logic translation tools implementing these mappings, off-the-shelf theorem provers for FOL have been turned into a practical reasoning systems for PMLs.  A reasoning tool based on this idea is MSPASS
~\citep{MSPASS}. Related approaches at generic theorem proving for different non-classical logics include the tableau-based theorem provers 
LoTReC~\citep{LoTREC}, 
MeTTeL~\citep{MetTeL} 
and the tableau workbench~\citep{TWB}. 
These systems allow the syntax and proof rules of the logic of interest to be explicitly specified in a respective interface from which they then generate a custom-tailored, tableau based theorem prover on the fly.\footnote{Further related systems and tools are described and linked online at \url{http://www.cs.man.ac.uk/~schmidt/tools/}} However, they are typically restricted to propositional non-classical logics only, which significantly limits their range of applications. In particular, non-trivial rational arguments in philosophy and metaphysics are clearly beyond their scope. Fact is: There are numerous reasoning tools available for PMLs, but only a handful implemented systems for first-order modal logics (FOML)~\citep{C34}. And, prior to the semantical embedding approach, there was not a single, practically available theorem prover for higher-order modal logics (HOML).

In the translation approach, which is generally not restricted to PMLs and FOL, the external transformation tool typically embodies and expands the semantics of the source (aka object) logic which it then translates into the target logic. The target logic is assumed to have equal or higher expressivity than the source logic, and the external transformation tool operates at an (extra-logical) meta-level in which a semantically justified bridge is established between the former and the latter. But do we actually need to segregate all these components? Why not realising the very same basic idea within one and the same logic framework, so that the source logic, the target logic and the meta-level are all ``living'' in the same space, and so that the \emph{logic transformations can themselves be explicitly specified and verified by logical means}? 

This question has inspired research on shallow semantical embeddings in HOL~\citep{B9,C32,J25,J21}, where the HOL meta-level is utilized to explicitly model the source and target logic, and the mapping between them. Moreover, in contrast to related work, the approach does not stop at the level of propositional non-classical logics, but rather puts an emphasis on first-order and higher-order quantified non-classical logics to render it amenable for more ambitious applications, including rational arguments in metaphysics, where e.g. higher-order modal logics play an important role.  The choice of HOL at the meta-level is thereby not by accident, but motivated as follows:


\textbf{(A)} For most logics in the logic zoo formal notions of semantics have been depicted based on set theoretical means. Examples are the Tarskian style semantics of classical predicate logic and the Kripke style semantics of modal logic. HOL which, thanks to its $\lambda$-notation, allows sets (e.g. $\{x \mid p(x)\}$) to be modeled by their corresponding characteristic functions (e.g. $\lambda x.  p(x)$), is well suited to elegantly encode many such set theoretic notions of semantics explicitly in form of a simple equational theory. 
The fact that HOL is sufficiently expressive is actually not so surprising when noting that \emph{an (informal) notion of classical higher-order logic is typically also the meta-logic of coice in most contemporary  logic or maths textbooks}. 

\textbf{(B)} \emph{Interactive theorem proving in HOL is already well supported in practice.} Powerful interactive provers have been developed over the past decades, including e.g. Isabelle/HOL~\citep{Isabelle}, 
HOL4~\citep{HOL}, 
HOL light~\citep{HOLlight} 
and 
PVS~\citep{PVS}. 
They often come with comfortable user-interfaces and intuitive user interaction support. Related proof assistants, which can also be turned into reasoners for HOL, include 
Coq~\citep{Coq}, 
Nuprl~\citep{Nuprl} 
and 
Lean~\citep{Lean}. 
Note that proof assistants have recently attracted lots of attention in mathematics, for example, in the context of Hales' successful verification of his proof of the Kepler conjecture\footnote{Johannes Kepler (1571-1630) stated the conjecture that the most dense way of stapling cannon balls (or oranges and alike)  is the form of a pyramid; the conjecture can be generalized beyond 3 dimensional space.}. While human experts alone had previously failed to fully assess his proof (this has happened for the first time in history) his formal verification attempt within the proof assistant HOL light succeeded \citep{KeplerCojectureProof}.\footnote{See also the following articles in New Scientist: \url{http://tinyurl.com/gvxzx42} and \url{http://tinyurl.com/jr8rdfq}.}  We will be facing an increasing number of analogous situations in the future: human and machine interactions will generate increasingly complex artefacts across all sciences, which, due to their sheer complexity and/or reasoning depth, will be deprived of traditional means of human assessment. We instead need new forms and means of scientific judgement, which again employ computer technology to overcome these challenges. However, ideally this computer technology is trusted (e.g. verified) and/or  delivers rational arguments back in a form amenable to human understanding and judgement.

\textbf{(C)} \emph{Also automated theorem proving in HOL has recently made significant progress.} Theorem provers such as LEO-II~\citep{LEO-II}, Satallax~\citep{Satallax} and the model finder Nitpick~\citep{Nitpick} have been successfully applied in a range of applications.  Moreover, new reasoning systems, such as the Leo-III prover~\citep{Leo-III} are currently under development.

\textbf{(D)} \emph{Interactive and automated reasoning in HOL has recently been well integrated.} Proof ``Hammering'' tools \citep{Hammers}, such as
Sledgehammer~\citep{Sledgehammer}
and 
Hol(y)Hammer~\citep{HolyHammer}
, are now available. They allow the interactive users of proof assistants such as  Isabelle/HOL and HOL light to conveniently call FOL and HOL reasoners in the background (even in parallel and remotely over the internet). Suitable logic transformations are realized within these systems and results are appropriately mapped back into trusted proofs the hosting proof assistants. Further projects have recently been funded in this area, including Matryoshka\footnote{\url{http://matryoshka.gforge.inria.fr}}
, AI4REASON\footnote{\url{http://ai4reason.org/}}
and SMART\footnote{\url{http://cordis.europa.eu/project/rcn/206472_en.html}}
. These projects, which (partly) integrate latest machine learning techniques, will significantly further improve proof automation of routine tasks in interactive proof assistants, with the effect that users can better concentrate on challenge aspects only.  



\paragraph{So, how does the semantical embedding approach work?}  Let L be an object logic of interest, for example, HOML as often required in metaphysics.  \emph{The overall idea is to provide a lean and elegant equational theory which interprets the syntactical constituents of logic L as terms of the target (and meta-)logic HOL.} Different to the traditional translation approach, this connection, i.e. the equational theory, is itself formalized in HOL. Moreover, in contrast to a deep logical embedding, where (the syntax and) the semantics of L would be formalized in full detail, only the crucial differences in the semantics of both are addressed in the equational theory and the commonalities, such as the notions of domains, are shared. Regarding the HOML  L and HOL, for example,  a crucial difference lies in the possible world semantics of L, and, hence, the equational theory provides an explicit modeling of this particular aspect of modal semantics.  More concretely, it associates the Boolean valued formulas $\varphi_o$ of L with world-predicates ($\lambda$-abstractions) $\varphi_{i\rightarrow o}$ in HOL (where $i$ stands for a reserved type for worlds). To establish such a mapping it essentially suffices to 
equate the logical connectives of L (e.g. $\wedge$ and $\nec$) with corresponding world-lifted predicates and relations in HOL (e.g. $\lambda \varphi. \lambda \psi. (\lambda w. \varphi w \wedge \psi w)$ and $\lambda \varphi. (\lambda w. \forall v. r(w,v) \rightarrow \varphi v$, where constant symbol $r$ denotes an accessibility relation between possible worlds). The mapping of constant symbols and variables of $L$ is then trivial, since only a type-lifting is required. Most importantly, the  mapping of L to HOL can be given in form of a finite set of quite simple equations (in fact, abbreviations); no explicit recursive definitions are required. 
Generally note the way in which the dependency of logic L on possible worlds is made explicit while other aspects and parameters of its semantic interpretation, such as the underlying semantic domains, remain (implicitly) shared between both logics.

\begin{figure}[t]\centering 
  \includegraphics[width=\columnwidth]{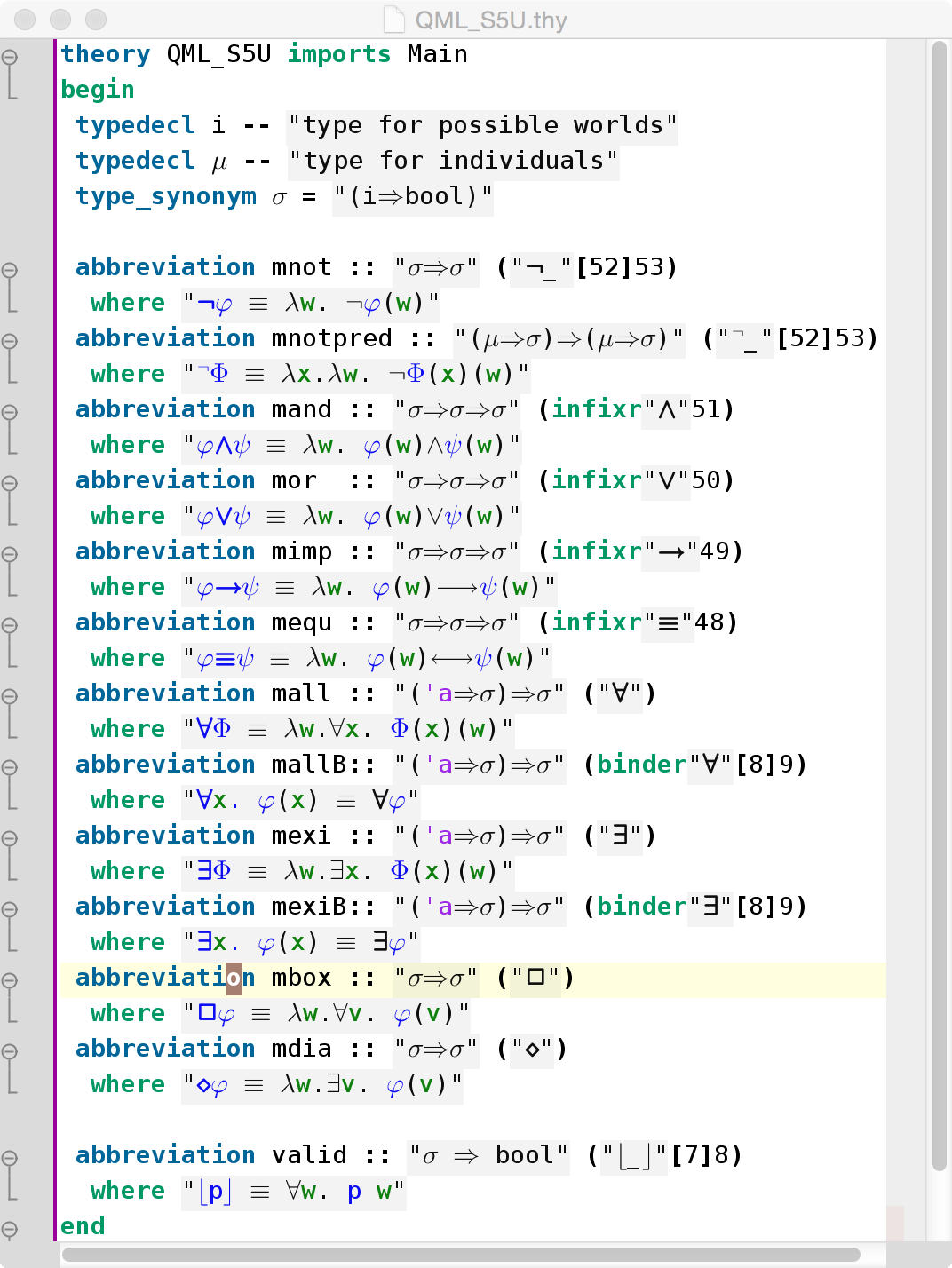}
\caption{A lean and elegant equational theory encoded in Isabelle/HOL which semantically embeds source logic HOML in target logic HOL; the equations are stated in meta-logic HOL. Here, a modal logic S5 with universal accessibility relation is exemplarily embedded. Moreover, type polymorphism is employed in the equations for the quantifiers. This way the otherwise required enumeration of quantifier-equations for all types can be avoided.}
\label{HOML}
\end{figure}

An interesting aspect is that \emph{the approach scales well even for first-order and higher-order quantifiers}. Thus, we can identify a fragment of HOL which, modulo the above sketched world-type-lifting, corresponds to HOML. This may seem astonishing, since HOML may appear more expressive than HOL at first sight.
Figure \ref{HOML} presents such an exemplary equational theory encoded in the proof assistant Isabelle/HOL. 
Formalisation tasks in challenging application areas (such as metaphysics) requiring HOML can now be carried out within Isabelle/HOL by using the HOML syntax as introduced. 
The HOL meta-logic guarantees global coherence and e.g. also enables for global consistency checks. And, modulo the embeddings in HOL, the automated reasoning tools available in Isabelle/HOL can now be reused.

Similar equational theories can be given for a wide range of non-classical logics (see e.g. the logics mentioned in \S\ref{applications} many of which have prominent applications in artificial intelligence, computer science, philosophy, maths and computational linguistics. Note that there is currently no other practically available approach in which a comparative range of logic embeddings has been established in practice. Moreover, soundness and completeness of the approach has already been established for a wide range of logics; thereby Henkin semantics is typically assumed for both HOL and the embedded source logics L (in case L goes beyond first-order).




\section{Some Exemplary Applications} \label{applications}
Obviously, \emph{the range of possible applications of the approach is very wide}. In fact, due to its generality, very few conceptual limitations are known at this point.\footnote{Eventually the use of HOL at the meta-level, as opposed to an even more expressive meta-logic, can be seen as conceptual limitation. However, there is no reason why HOL could not be exchanged by an even more expressive meta-logic, provided that practical reasoning tools are available for it.}  Some exemplary  application directions, which have already been addressed in pilot studies, are outlined.  From a practical perspective a relevant question clearly is whether the theorem provers performance scales beyond small proof of concept examples. This question has to be assessed individually for each application domain. However, the experience from the pilot studies mentioned below is that \emph{the approach indeed matches and may even outperform human reasoning capabilities} in individual application domains (e.g. flaws in human refereed research papers and textbooks have been revealed). Another reassuring fact is that in particular  in the area of metaphysics the argumentation granularity (size of single argumentation steps) that was supported in full automatic mode by the approach well \emph{matched the typical argumentation granularity in human generated, rational (masterpiece) arguments}.  Moreover, note that only the propositional fragments (and in a few cases the first-order fragments) of the logics mentioned below have been automated in practice before. \emph{The semantical embedding approach, however, scales for their propositional, first-order and even higher-order logic fragments.}
Future work includes the widening of the range of application pilot studies, in particular, towards the modeling and assessment of rational arguments between intelligent machines and humans. It can be expected that, in the long-run,  the combination of expressive quantified non-classical logics will become highly relevant in this context.

\subsection{Philosophy}

\paragraph{Masterpiece Rational Arguments in Metaphysics.}
Numerous modern variants of the \emph{Ontological Argument} for
the existence of God, one of the still vividly debated masterpiece
arguments\footnote{See e.g.~\citet{sobel2004logic} for
  more details on the ontological argument.} in metaphysics, have been rigorously analysed on the computer. In the course of these experiments, the higher-order ATP LEO-II \citep{J30} detected an (previously unknown!) inconsistency in Kurt \citeauthor{GoedelNotes}'s (\citeyear{GoedelNotes}) prominent, higher-order modal logic variant of the argument, while Dana \citeauthor{ScottNotes}'s (\citeyear{ScottNotes}) slightly different variant of the argument was completely verified in the interactive proof assistants Isabelle/HOL and Coq.  Further relevant insights contributed or confirmed by ATPs e.g. include the separation of relevant from irrelevant axioms, the determination of mandatory properties of modalities, and undesired side-implications of the axioms such as the ``modal collapse''\footnote{The modal collapse is a sort of constricted
  inconsistency at the level of possible world semantics. The
  assumption that there may actually be more than one possible world
  is refuted; this follows from G\"odel's axioms as the ATPs
  quickly confirm. In other words, G\"odel's axioms, as a side-effect, imply
  that everything is determined (we may even say: that there is no
  free will).}. 
 The main results about G\"odel's and Scott's proofs have been presented at ECAI and IJCAI conferences \citep{C40,C55}.

Further variants of G\"odel's axioms were proposed by Anderson, Bjordal and H\'ajek \citep{Anderson,AndersonGettings,Hajek1,Hajek2,Hajek3,Bjordal}. These variants have also been formally analysed, and, in the course of this work, theorem provers have even contributed to the clarification of an unsettled philosophical dispute between Anderson and H\'ajek~\citep{J32}.  Moreover, the modal collapse, whose avoidance has been the key motivation for the contributions of Anderson, Bjordal and H\'ajek (and many others), has been further investigated~\citep{B15}. 
Several further contributions complete these initial experiments on the formal assessment of rational arguments in metaphysics \citep{J28,C44,C52,C50,C46,C42,W55,W50}.

\paragraph{Principia Metaphysica.}
Analyzing masterpiece rational arguments in philosophy with the semantical embedding approach on the computer is not trivial. However, it still leads to comparably small corpora of axioms, lemmata and theorems, and it does thus not yet provide feedback on the scalability of the approach for larger and more ambitious projects. For that reason another challenge has recently been tackled: the \emph{Principia Logico-Metaphysica} (PLM) by \citet{zalta16:_princ_logic_metap}. The PLM is intended to provide a rigorous formal basis for all of metaphysics and the sciences; this includes a (flexible) foundation for mathematics and in this sense it is more ambitious than Russel's Principia Mathematica.
Since Zalta has chosen a hyperintensional (relational) higher-order modal logic S5 as the logical foundation of his PLM, it has hence been an open challenge question whether this very specific logical setting can still be suitably encoded in the semantical embedding approach. Besides hyperintensionality, a particular challenge concerns the conceptional gap between the relational and functional bases of the logic of the PLM and HOL, which imply different strengths of comprehension principles, which in turn are of significant impact to the entire theory (full comprehension in the PLM causes paradoxes and inconsistencies,~\citet{DBLP:journals/logcom/OppenheimerZ11}). 

Despite these challenges, the ongoing work on the PLM has progressed very promisingly. In fact, most of the PLM has meanwhile been represented and partially automated in Isabelle/HOL by using the semantical embedding approach.\footnote{See \url{https://github.com/ekpyron/TAO}, respectively ~\url{https://github.com/ekpyron/TAO/blob/master/output/document.pdf}}

\paragraph{Other Logics in Philosophy.}
The approach has recently been successfully applied to other prominent logics in philosophy, including quantified conditional logics \citep{J31,C37}, multi-valued logics~\citep{J33} and paraconsistent Logics~\cite[Sec.~5.4]{C46}.

\paragraph{Award Winning Lecture Course}
The successes presented above and below have inspired the design of a worldwide new lecture course on \emph{Computational Metaphysics} at FU Berlin \citep{C58}.\footnote{The lecture course, held in summer 2016, has received FU Berlin's central teaching award; see \url{http://www.fu-berlin.de/campusleben/lernen-und-lehren/2016/160428-lehrpreis/index.html}. 
}  In this course the above research on the formalisation of ontological arguments and the foundations of metaphysics led into a range of further formalisation projects in philosophy, maths and computer science.  Some of the student projects conducted in this course have resulted in impressive new contributions. For example, a computer-assisted reconstruction of an ontological argument by Leibniz will appear as a chapter in a book dedicated to the 300th anniversary of Leibniz's death \citep{B16}. 
Also core parts of the textbooks by \citet{fitting02:_types_tableaus_god} and \citet{boolos93:_logic_provab} have meanwhile been formalised.\footnote{The sources of the formalisation of Fitting's work are available at \url{https://github.com/cbenzmueller/TypesTableauxAndGoedelsGod}.}
A key factor in the successful implementation of the course has been, that a single methodology and overall technique (the semantical embedding approach) was used throughout, enabling the students to quickly adopt a wide range of different logic variants  in short time within a single proof assistant framework (Isabelle/HOL). The course concept is potentially suited to significantly improve interdisciplinary, university level logic education.



\subsection{Mathematics}
\paragraph{Free Logics.}
Prominent, open challenges for formalisation in mathematics (and beyond) include the handling of partiality and definite descriptions.  \emph{Free logic} \citep{lambert12:_free_logic,Scott} adapts classical logic in a way particularly suited for handling such challenges. Free logics have interesting applications, e.g. in natural language processing and as a logic of fiction. In mathematics, free logics are particularly suited in applications domains such as category theory or projective geometry (e.g. morphism composition in category theory is a partial operation). Similar to the other non-classical logics mentioned before, free logics can be elegantly embedded in HOL \citep{C57}.

\paragraph{Category Theory.}
Utilizing this embedding of Scott's (\citeyear{Scott}) approach to free logic in HOL, a systematic theory development in category theory has recently been contributed. In this 
exemplarily study six different but closely related axiom systems for category theory have been formalized in Isabelle/HOL and proven mutually equivalent with automated theorem provers via Sledgehammer. In the course of these experiments, the provers revealed a technical flaw (constricted inconsistency or missing axioms) in the well known category theory textbook by \citet{FreydScedrov}.

\subsection{Artificial Intelligence and Computer Science}
Most of the above mentioned logics (and respective experiments) are obviously relevant also for applications in artificial intelligence and computer science.  Further relevant experiments include:

\paragraph{Epistemic and  Doxastic Logics.} Epistemic logic supports e.g. the modeling of knowledge of rational agents. Doxastic logic is about the modeling of agent beliefs. Both are just particular multi-modal logics and thus amenable to the semantical embedding approach. Respective experiments show that the approach indeed works well for elegantly solving prominent puzzles about knowledge and belief in artificial intelligence \citep{J25,C61}, including the well known wise men puzzle resp. muddy children puzzle.

\paragraph{Time and Space.} The reasoning about time and space has been a long standing challenge in artificial intelligence, in particular, when combined reasoning about time, space and eventually further modal concepts is required. Again, the semantical embedding approach can provide a possible solution, see e.g. the combination of spatial and epistemic reasoning outlined in \citep[Sec. 6]{J25}.

\paragraph{Description Logics.}  Description logics are prominent e.g. in the semantic web community. However, description are basically just a reinvention
multi-modal logics (the base description logic ALC corresponds to a basic multi-modal logic K), and thus the semantical embedding approach elegantly applies. Hence, the shallow embedding approach applies also to a range of prominent description logics, and the mentioned logic correspondences can even be verified in it.

\paragraph{Many-valued Logics.}
Many-valued logics have applications, for example, in philosophy, mathematics and computer science. Theorem provers for various propositional, first-order and higher-order many-valued logics can easily be obtained by utilising the semantical embedding approach. An exemplary semantical embedding of the multi-valued logic SIXTEEN has been provided in~\citep{J33}.

\paragraph{Access Control Logics (Security).}
The semantical embedding approach also applies to security logics, and respective experiments for access control logics have been reported \citep{C27}.

\section{Summary and Outlook}
The semantical embedding approach, which utilises classical higher-order logic at meta-level to encode (combinations of) a wide range of non-classical logics, has many applications e.g. in artificial intelligence, computer science, philosophy, mathematics and (deep) natural language processing.  Automation of reasoning in these logics (and their combinations) is achieved indirectly with off-the-shelf reasoning tools as currently developed, integrated and deployed in modern higher-order proof assistants. The range of possible applications of this universal reasoning approach is far reaching and, as has been demonstrated, even scales for non-trivial rational arguments, including masterpiece arguments in philosophy. 

A relevant and challenging future application direction concerns the application of the semantical embedding approach for the modeling of ethical, legal, social and cultural norms in intelligent machines, ideally in combination with the realisation of human-intuitive forms of rational arguments in machines complementing internal decision making means at the level of statistical information and subsymbolic representations. To enable such applications, the author is currently adapting the semantic embedding to cover also recent works in the area of deontic logics (such  as~\cite{DBLP:journals/jphil/MakinsonT00} and~\cite{DBLP:journals/logcom/CarmoJ13}).



\subsection*{Acknowledgements:} This work has been supported by the following research grants of the German Research Foundation DFG: BE 2501/9-2 (Towards Computational Metaphysics) and BE 2501/11-1 (Leo-III). I cordially thank all my collaborators of these and other related projects.  This includes (in alphabetical order): Larry Paulson, Dana Scott, Geoff Sutcliffe, Alexander Steen, Max Wisniewski,  Bruno Woltzenlogel-Paleo and Edward Zalta.

\small
\bibliographystyle{named}

\end{document}